\documentclass{article}
\usepackage{spconf,amsmath,epsfig}
\newcommand{\savespacebeforetable}{\vspace{-10pt}}

\title{Reranking Machine Translation Hypotheses With Structured and Web-based Language Models}
\name{{\em Wen Wang, Andreas Stolcke, Jing Zheng}}

\address{Speech Technology and Research Laboratory\\
SRI International, Menlo Park, CA, USA \\
{\small \tt \{wwang,stolcke,zj\}@speech.sri.com}}
\begin{document}
\ninept
\maketitle
\begin{abstract}
In this paper, we investigate the use of linguistically motivated and
computationally efficient structured language models for reranking
N-best hypotheses in a statistical machine translation system. These
language models, developed from Constraint Dependency Grammar parses,
tightly integrate knowledge of words, morphological and lexical
features, and syntactic dependency constraints. Two structured
language models are applied for N-best rescoring, one is an
almost-parsing language model, and the other utilizes more syntactic
features by explicitly modeling syntactic dependencies between
words. We also investigate effective and efficient language modeling
methods to use N-grams extracted from up to 1 teraword of web
documents. We apply all these language models for N-best re-ranking on the NIST and DARPA GALE program\footnote{The goal of the GALE program is to develop computer
software techniques to analyze, interpret, and distill information from
speech and text in multiple languages. For processing languages other than English, machine translation is an important module in the pipeline.}
2006 and 2007 machine translation evaluation tasks and find that
the combination of these language models increases the BLEU score up to 1.6\% absolutely
on blind test sets.
\end{abstract}
\vspace{-0.1in}
\section{Introduction}
\label{sec:intro}
\vspace{-0.05in}
The goal of statistical machine translation (SMT) is to find the best translation $\hat{e}_{1}^{\hat{I}}=\hat{e}_{1} \ldots \hat{e}_{i} \ldots \hat{e}_{\hat{I}}$ of source language sentence
$f_{1}^{J}=f_{1} \ldots f_j \ldots f_{J}$ where
{\scriptsize
\begin{eqnarray*}
\hat{e}_{1}^{\hat{I}}  & = & \arg\max_{I, e_{1}^{I}}{Pr(e_{1}^{I}|f_{1}^{J})} \\
			& = & \arg\max_{I, e_{1}^{I}}{Pr(f_{1}^{J}|e_{1}^{I}) \cdot Pr(e_{1}^{I})}
\end{eqnarray*}}
\noindent
Instead of using this source-channel approach, the direct modeling of the posterior probability $Pr(e_{1}^{I}|f_{1}^{J})$ can be computed as follows by using a log-linear model \cite{Och:acl2002}:
{\scriptsize
\begin{eqnarray*}
Pr(e_{1}^{I}|f_{1}^{J}) = \frac{exp(\sum_{m=1}^{M}\lambda_{m}h_m(e_{1}^{I},f_{1}^{J}))}{\sum_{{e^{'}}_{1}^{I^{'}}}exp(\sum_{m=1}^{M}\lambda_{m}h_m({e^{'}}_{1}^{I^{'}},f_{1}^{J}))}
\end{eqnarray*}}
\noindent
where $\lambda_m$'s are the weights (or scaling factors) for the models denoted by feature functions $h_m(\cdot)$. Since we can ignore the denominator, which is a normalization factor and is constant for a source sentence $f_{1}^{J}$, the goal of translation is to find
\begin{eqnarray*}
\hat{e}_{1}^{\hat{I}} = \arg\max_{I, e_{1}^{I}}{\sum_{m=1}^{M}\lambda_m h_m(e_{1}^{I}, f_{1}^{J})}
\end{eqnarray*}
\noindent
With the above approach, we can easily integrate additional models
$h(\cdot)$ as new knowledge sources and train the weights either using
the maximum entropy principle or optimizing them based on the final
translation performance using a certain evaluation metric such as BLEU or word error rate (WER). In this
paper, we will investigate the efficacy of adding various language model (LM) reranking
scores as additional knowledge sources and optimize the weights using
minimum error training.

There has been much effort recently in MT on adding syntactically motivated features. 
Och and others \cite{Och:naacl2004} investigated the efficacy of integrating syntactic
structures into a state-of-the-art SMT 
system by introducing feature functions representing syntactic
information and discriminatively training scaling factors on a set of
development N-best lists. They obtained consistent and significant
improvement from the implicit syntactic features produced by IBM model
1 scores, but rather small improvement from other syntactic
features, ranging from shallow to deep parsing approaches.
Recently, Hasan and others \cite{Hasan:eacl2006} observed promising improvement of MT performance in a reranking framework by using supertagging and lightweight dependency analysis, a link grammar parser, and a maximum entropy based chunk parser.
They achieved up to 0.7\% absolute increase on BLEU on C-Star'03 and IWSLT'04 tasks. 
In this paper, we investigate the efficacy of structured LMs, which integrate lexical features and syntactic constraints, in an N-best reranking framework for SMT.

We also explore the use of large LMs derived from world-wide-web data
for SMT reranking.
Prior work along these lines includes distributed language modeling for N-best reranking \cite{Zhang:emnlp2006} as well as introducing a simple, inexpensive smoothing method that can work reasonably well on very large amounts of data \cite{Brants:emnlp2007}.
In this paper we make use of efficient deleted interpolation smoothing to 
accommodate very large databases, and address the problem of modified Kneser-Ney (KN)
smoothing \cite{chen98} from incomplete N-gram distributions.
The latter is required because most of our web data is provided in the form
of N-gram corpora, without access to the raw data.

\vspace{-0.1in}
\section{MT system description} 
\label{sec:system} 
\vspace{-0.05in}
SRI's 2007 GALE evaluation Arabic-to-English and Chinese-to-English translation systems
consist of two passes of decoding. The first pass uses a
hierarchical phrase decoder developed at SRI to perform integrated decoding
with a standard 4-gram LM to generate N-best
lists. The basic phrases and hierarchical rules were extracted from
parallel corpora and word-alignments provided by RWTH Aachen University,
similar to David Chiang's approach
\cite{Chiang:acl2005}. The second pass rescores the N-best lists using
several LMs trained on different corpora and estimated in
different ways. The scores are then combined in the log-linear
modeling framework \cite{Och:acl2002} along with other features used
in the SMT system, including rule probabilities $p(f|e)$, $p(e|f)$,
lexical weights $pw(f|e)$, $pw(e|f)$\cite{Koehn:naacl2003},  sentence
length, and rule counts. We optimized the weights using the minimum
error training method to maximize BLEU scores using Amoeba simplex
search on N-best lists, which could easily be extended to other
objective functions such as word error rate (WER) and translation
error rate (TER).  The weights were optimized on a development set
(GALE dev07) and applied to the blind test set, NIST MT eval06 GALE portion (denoted eval06).

\vspace{-0.1in}
\section{Structured LMs}
\label{sec:structured-lm}
\vspace{-0.05in}
Syntax-based translation models have been shown to capture better
long-range word order difference between the source and target language,
and to produce higher quality translations than the standard phrase-based
models\cite{Chiang:acl2005,Marcu:emnlp2006}. However, in the language modeling aspect, even the
state-of-the-art syntax-based systems are not yet able to utilize
syntactic properties of languages; instead, they rely on the standard n-gram language
models that capture only local features.

In this paper, we investigate the idea of using relatively loose
coupling methods to access rich syntactic information for MT compared
to syntax-based models, by using structured LMs for N-best
reranking. In the area of automatic speech recognition (ASR),
structured LMs have recently been shown to give
significant improvement in recognition accuracy relative to
traditional word n-gram models \cite{Chelba:98,Roark:2001}.
In \cite{emnlp2002},  we developed an
almost-parsing, SuperARV-based language model within the Constraint
Dependency Grammar (CDG)
framework for speech recognition. In this work  we extend the almost-parsing LM to MT N-best reranking applications and also investigate the use of a parser LM for reranking. Note that it is important for these structured LMs to be computationally efficient
in order to be applied to N-best rescoring for MT tasks, which in general deals with large amounts of data.

\vspace{-0.05in}
\subsection{Almost-parsing language model}
\label{sec:sarv-lm}
\vspace{-0.05in}
\subsubsection{Summary of the model}
\vspace{-0.05in}
The SuperARV LM \cite{emnlp2002} is a highly lexicalized probabilistic
LM based on Constraint Dependency Grammars (CDGs), with grammar rules
factored at the word level.  It tightly integrates multiple
knowledge sources, for example, word identity, morphological features,
lexical features that have synergy with syntactic analyses (e.g., gap
propagation, mood),  and syntactic and semantic constraints, at both
the {\em knowledge representation level\/} and {\em model level}.

Knowledge representation level integration was achieved by
introducing a linguistic structure, called a super abstract role value
({\em SuperARV}), to encode multiple knowledge sources in a uniform
representation that is much more fine-grained than part-of-speech
(POS). A SuperARV is an abstraction of the joint assignment of all
dependencies for a word, 
formally defined as a four-tuple
$\langle C, F$, $(R, L, UC, MC)+, DC \rangle$, where C is the
lexical category of the word, $F=\{Fname_{1}$ $=Fvalue_{1}$,
$\ldots,Fname_{f}=Fvalue_{f}\}$ is a feature vector ($Fname_{i}$ is
the name of a feature and $Fvalue_{i}$ is its corresponding value),
(R, L, UC, MC)+ is a list of one or more four-tuples, each
representing an abstraction of a role value assignment, where $R$ is a
role variable (e.g., governor), $L$ is a functionality label (e.g.,
np), $UC$ represents the relative position relation of a word and its
dependent (i.e., modifiee), $MC$ is the lexical category of the
modifiee for this dependency relation, and $DC$ represents the
relative ordering of the positions of a word and all of its modifiees.
Hence, the SuperARV structure for a word provides 
admissibility constraints on
syntactic and lexical environments in which it may be used. 

The second type, model-level integration was accomplished by jointly
estimating the probabilities of a sequence of words $w_{1}^{N}$ and
their SuperARV membership $t_{1}^{N}$:
{
\begin{eqnarray*} P(w_{1}^{N}t_{1}^{N}) & = &
\prod_{i=1}^{N}P(w_{i}t_{i} | w_{1}^{i-1}t_{1}^{i-1}) \\
			        & = & \prod_{i=1}^{N}P(t_{i} |
w_{1}^{i-1}t_{1}^{i-1}) \cdot P(w_{i} | w_{1}^{i-1}t_{1}^{i})
\label{eqno2} 
\end{eqnarray*}} 
\noindent
We use this to enable the joint
prediction of words and their SuperARVs so that word identity
information is tightly integrated at the model level.  The SuperARV LM
is fundamentally a class-based LM using SuperARVs as classes. N-gram
conditional probabilities are estimated as follows:
{
\begin{eqnarray*} 
\label{entropy}
P(w_i | w_{1}^{i-1}) = \frac{\sum_{t_{1,i}}P(w_it_i |
w_{1}^{i-1}t_{1}^{i-1})P(w_{1}^{i-1}t_{1}^{i-1})}{\sum_{t_{1,i-1}}P(w_{1}^{i-1}t_{1}^{i-1})}
\nonumber 
\end{eqnarray*}} 
\noindent
Detail on resolving parameter estimation issues appear in \cite{emnlp2002}. The SuperARV language model
generates almost-parses for input sentences,  since it does not
explicitly selects modifiees but simply determines directions and position relations of all dependency links for each word. The SuperARV LM is most closely related to the
almost-parsing LM developed by Srinivas
\cite{srinivas-thesis} based on the supertags that are the elementary
structures of Lexicalized Tree-Adjoining Grammar. 
For training SuperARV LMs, we developed a methodology to automatically transform
context-free grammar (CFG) constituent bracketing into CDG annotations
\cite{wangphd}. Then SuperARVs can be extracted from CDG parses and word and SuperARV statistics can be estimated.

\vspace{-0.05in}
\subsubsection{Modeling numbers and punctuation}
\vspace{-0.05in}
When developing an almost-parsing language model for SMT N-best reranking, we modified the modeling of the SuperARV LM
in two ways. First, to improve generalization and coverage, numbers are
mapped to a macro word ``\$number'' during preprocessing on the parallel
data and other target language model training data. For training
the structured language models in this work, we delayed this
mapping procedure by first generating valid CFG parse trees on the
original word formats of numbers, then mapping the numbers in the parse trees
to ``\$number''.

Second, unlike in ASR language modeling, punctuation is present
in the MT N-best hypotheses. Punctuation provides important syntactic information for SMT and needs to be modeled.
To this end, we categorize
punctuation marks into sentence-final punctuation (i.e., period, question
mark, exclamation) and intra-sentence punctuation, and discriminate
between them in modeling dependencies between punctuation and word
tokens. When generating the CDG annotations for punctuation from CFG
parse trees, we use their surface format, such as  ``.'' and ``,'', as
their lexical categories\footnote{We found that this fine-grained
categorization produced better
parsing performance compared to clustering all punctuation into
a single class.}. Punctuation tokens bear no lexical
features, i.e., no $F$ is defined for them in the SuperARV tuple. When defining
dependency relations with other words, as is common in the parsing
community \cite{collins96}, we attach punctuation marks as high as possible
in the CFG parse trees.  We select the root of a sentence to be the
headword for sentence-final punctuation marks. For intra-sentence
punctuation marks, we treat them similarly to coordination by defining the
headword of the following phrase as the headword of the punctuation mark.
When a sentence is incomplete and an intra-sentence
punctuation mark appears at the end of a phrase, instead of between phrases,
we pick up the headword of the preceding phrase as the headword of this
punctuation.

\vspace{-0.05in}
\subsection{Parser LM}
\label{sec:parser-lm}
\vspace{-0.05in}
In \cite{Wang:asru2003}, we developed a statistical full parser-based
LM. Given a sentence $W=w_1 \ldots w_i \ldots w_n$,  the LM combines
SuperARV tagging and modifiee specification by predicting the SuperARV
tag sequence $S$ and the set of dependency relations $D$ for
$S$ (either in a loose or a tight coupling scheme). Compared to the almost-parsing LM, this language model can
utilize long-distance dependency constraints and subcategorization
information for word predictions. We observed improvement on word
prediction accuracy by strengthening syntactic constraints as
reported on other structured language models \cite{Xu02}. However,
this full parser based LM utilizes statistics of dependencies between
all pairs of words, and hence the computational complexity is quite high. In this section, we present a new parser LM that is much more efficient.

\vspace{-0.05in}
\subsubsection{Using the baseNP model} 
\vspace{-0.05in}
First, we explored the idea of
using baseNPs in dependency descriptions for parses
\cite{collins96}. Given a sentence $W$, we generate the reduced
sentence $\bar{W}$ for it by first marking all baseNPs and then
reducing all baseNPs to their headwords. A baseNP (or minimal NP)
is a non-recursive NP such that none of its child constituents are NPs
\cite{collins96}. For example, a sentence ``Mr. Viken is chairman of
the Elsevier N.V., the Dutch publishing group'' will be reduced to
``Viken is chairman of N.V., group''. Note that words internal to
baseNPs cannot modify words outside their baseNP so they do not
contribute much to enforcing long-span dependency constraints for the
parser LM.  In this way, words internal to baseNPs are not used for
training and computing dependency statistics (during rescoring), and hence
the efficiency of the LM is significantly improved.

To identify baseNPs for an input sentence $W$, we apply the baseNP
model which is essentially a tagger to tag the boundaries between words
using tags from the set ${S, C, E, B, N}$ (denoting whether the boundary is at the
start of a baseNP, continues a baseNP, is at the end of a baseNP, is
between two adjacent baseNPs, or is between two words neither of which
belongs to any baseNPs, respectively). Given the gap between words
$w_{i-1}$ and $w_i$ denoted $G_i$, similar to \cite{collins96}, we use
the two words to the left and right of $G_i$ and their POS tags for
predicting the tag of $G_i$. Different from \cite{collins96}, instead
of just using commas as baseNP delimiters, we consider whether there
is an intra-sentence punctuation between the words by introducing the
variable $c_i$ ($c_i=1$ when there is an intra-sentence punctuation mark between $w_{i-1}$ and $w_i$,
and $c_i=0$ otherwise).  The baseNP model estimates the probability of
tagging baseNPs $B$ for the sentence $W$ as (define $P(G_1 | w_1, t_1)
= 1$):
\begin{eqnarray*}
P(B|W) = \prod_{i=1,n}P(G_i | w_{i-1}, t_{i-1}, w_i, t_i, c_i)
\end{eqnarray*}
where $t_i$ is the part-of-speech tag of word $w_i$.
The probability estimation is smoothed similar to \cite{collins96},
since that method proved to be simple and effective.

\vspace{-0.05in}
\subsubsection{Further simplification of the parser LM}
\vspace{-0.05in}
Our full parser LM \cite{Wang:asru2003} is a probabilistic generative
model. After introducing the baseNP modeling, we modified the full
parser LM as follows. Given $T$ as a dependency annotation for a
sentence $W$, the probability of $W$ based on this LM is computed as:
\begin{eqnarray*}
P(W) &=& \sum_{T} P(W, T) = \sum_{T} P(W, S, B, D) \\
     &=& \sum_{T} P(D | B, S, W) P(B|S,W) P(S,W)
\end{eqnarray*}
\noindent
With Viterbi approximations and making independence assumptions, we approximate $P(W)$ as:
\begin{eqnarray*}
P(W) \approx \hat{P}(D|\bar{S}) \hat{P}(B|W) \hat{P}(S,W)
\end{eqnarray*}
\noindent
This means that for an input sentence $W$, we first search
for the best SuperARV sequence and best baseNP sequence for it. Then based
on the reduced sentence $\bar{W}$ and the corresponding SuperARV tags
$\bar{S}$ for the reduced sentence, we search for the best dependency
relations that maximizes $P(D|\bar{S})$ using stack decoding as described in
\cite{Wang:asru2003}. Note that in practice this approximation should
be reasonable since by examining a non-trivial numbers of examples, we
found that most of the baseNPs are very well defined and that parses
among the highest scoring parses for a sentence generally share
identical or very similar baseNP sequences, which are either the best
baseNP sequence or very close to the best baseNP sequence.

\vspace{-0.1in}
\section{LMs for large web data collections}
\label{sec:google-yahoo-lm}
\vspace{-0.05in}
We trained two LMs that were based on N-gram corpora released by
Google and Yahoo, respectively. The Google LM was based on 5-grams
extracted from 1 terawords of web documents covering a wide range of
sources. To limit the size of the distribution, only N-grams occurring
40 times or more were included; still it contained 0.31G bigrams,
0.98G trigrams, 1.3G 4-grams and 1.2G 5-grams, over a vocabulary of 11M
words, and occupied about 25G on disk even after file compression.
To build an LM that avoids memory issues we implemented a count-based
LM representation in SRILM \cite{srilm:icslp2002} using the Jelinek-Mercer
deleted interpolation smoothing approach \cite{chen98}.
Interpolation weights were estimated on the held-out tuning set.
During weight estimation and testing, the model needs access to only
those N-grams occurring in the respective data sets, allowing for fast operation
with limited memory.  We refer to this type of LMs as a ``count-LM''.

The second web-based LM was trained on 5-grams provided by Yahoo,
extracted from about 3.4G words of news sources during a 1-year period
prior to the epoch of the GALE 2007 MT evaluation data.
Although containing all N-grams
occurring more than once, this collection was much smaller, but also
more recent and directly focused on the target domain (broadcast news)
than the Google corpus. It comprised about 54M bigrams, 187M trigrams,
288M 4-grams, and 296M 5-grams.  The Yahoo LM we trained was a
standard backoff LM using modified KN smoothing \cite{chen98},
for both 4-gram and 5-gram versions.

\begin{figure}
\caption{{\small Plot of count-of-count frequencies $F(k)$ according to the function
$1 \over { \log{F(k)} - \log{F(k+1)} }$.
The reciprocal of the slope of the graph gives $\alpha$ from Equation
\ref{eq:kncounts} and allows extrapolation to unknown values of $F(k)$.
In this case $\alpha = 2.42$ (4-grams) and $2.67$ (5-grams) on Yahoo N-grams.}}
\label{fig:kncounts}
\begin{center}
\includegraphics[width=0.7\columnwidth]{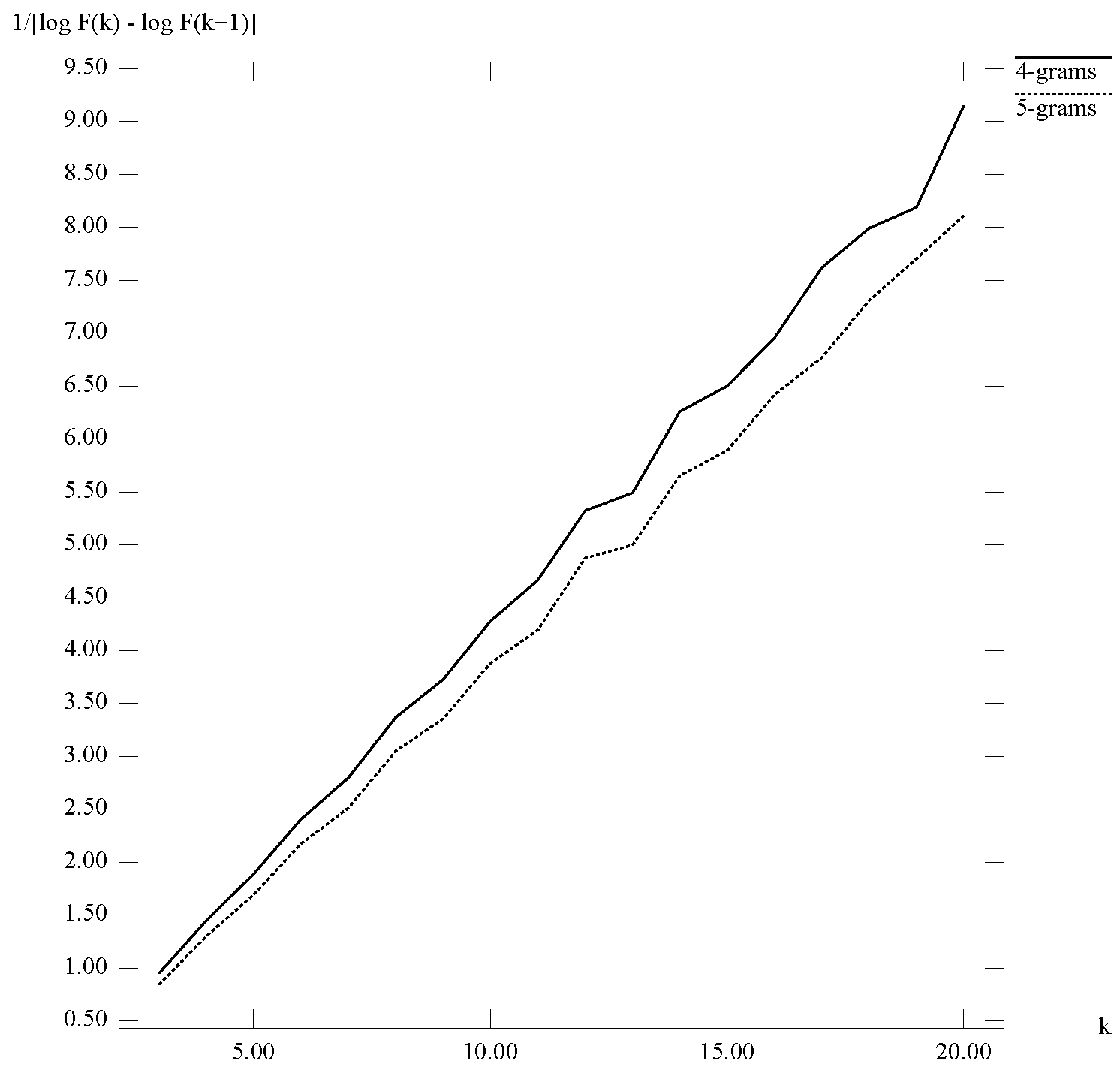}
\end{center}
\end{figure}

One interesting issue arising in building the Yahoo LM was that the
number of singleton N-grams was not available in Yahoo's release
(probably for the purpose of reducing the number of DVDs to store the data),
and yet is required to
compute the KN discounting values. By studying the
distribution of N-grams in various corpora for different languages, we
found an empirical law that seems to govern the progression of these
counts of counts in large natural data sets:
\begin{equation}
\log{F(k)} - \log{F(k+1)} = \frac{\alpha}{k}
\label{eq:kncounts}
\end{equation}
\newpage\noindent
where $F(k)$ is the number of distinct N-grams of count $k$, and $\alpha$ is a constant dependent on the corpus and the
N-gram order.
Figure~\ref{fig:kncounts} demonstrates the regularity of the data by
plotting the associated linear function.
This law allows us to extrapolate from the
available $F(k)$ to the missing ones, in this case,  $F(1)$,  by
estimating the factor $\alpha$ based on the available counts of counts.
After the extrapolation, we applied modified KN smoothing to train
4-gram and 5-gram LMs from the Yahoo data.

\vspace{-0.1in}
\section{Experiments and results}
\label{sec:experiments}
\vspace{-0.05in}
The DARPA GALE program machine translation evaluation includes testing
translation of text and machine transcription of recorded speech. The
test includes language data from both Arabic and Chinese. The input to
text MT is a variety of unstructured source language documents
taken from newswire publications (denoted {\bf NW}) and web-based
newsgroups (denoted {\bf WT}).   The input to the recorded speech
includes broadcast news (denoted {\bf BN}) and broadcast conversations (talk and call-in shows, denoted {\bf BC}).  In the following experiments,
we will investigate SMT performance on all of these four genres, NW, WT,
BN, and BC, for both Arabic-English and Chinese-English.

\vspace{-0.05in}
\subsection{Data}
\vspace{-0.05in}
The word 4-gram LM used for search in the MT system was trained on
various text sources with close to 5 billion tokens in total. We clustered them
into the following 4 categories: 
\begin{description}
\item[1:]  the English side of Arabic-English and Chinese-English parallel
data provided by LDC (270 million tokens for CE and 280 million tokens
for AE);
\item[2:]  all of the English BN and BC transcriptions, web text, and
translations for Mandarin and Arabic BN and BC, released under the
DARPA EARS and GALE programs
\footnote{http://projects.ldc.upenn.edu/gale/data/catalog.html lists
the LDC released corpora related to GALE.} (260 million tokens);
\item[3:] English Gigaword corpus (LDC2005T12, 2.5G tokens), North
American News text corpora (LDC95T21 and LDC98T30, 620 million
tokens);
\item[4:]  Web data collected by SRI and BBN. SRI collected 30 million
tokens of English news articles from June
2005 to January 2006. BBN's web data was collected by downloading
articles from news web sites that
    offer free access to their archives, with total word counts as
about 800 million tokens, and by crawling the ``internet archive'' 
and downloading past copies of pages from news websites,
with total word counts as about 300 million tokens.
\end{description}

All training text was preprocessed following RWTH Aachen's
treebank-style tokenization and text normalization toolkit, with some
bug fixes and further cleanup especially for acoustic transcripts.
The vocabulary of the word N-gram and structured language models is 2.3
million words, and we made sure all words appearing in the English side
of the parallel data are covered.  We trained separate word 4-gram LMs
using modified KN smoothing on each of the four sources, and then
created the final mixture LM by optimizing linear interpolation weights
of these component LMs on a heldout tuning set. Note that we used a
very high cutoff when training the component LM for the BBN web data,
 since they appear to be much more noisy compared to the LDC released
news text. To compensate for this, we also trained a count-LM using
all BBN web data without cutoff, denoting this LM as BBN-web-lm in the
following experiments.

For the two structured language models, due to limited time for system
development, we used only the first two sources for training the
almost-parsing LM and the second source only for training the parser LM.

For N-best reranking, we used the word 4-gram LM (denoted
{\bf 4-gram} or
 {\bf 4g}) (i.e., scores from search are used in combination), the almost-parsing LM (denoted {\bf sarv}), the parser LM
(denoted {\bf plm}), the 5-gram count-LM trained on the Google
N-grams (denoted {\bf google}), the modified Kneser-Ney smoothed 4-gram
LM trained on the Yahoo N-grams (denoted {\bf yahoo}), and the
5-gram count-LM trained on all BBN web data (denoted {\bf
wlm}). Note that the perplexities from different language models using
different vocabularies are not directly comparable, so here we only
compared the perplexities from 4g, yahoo, sarv, and plm, on a 32K
words LM tuning set. The perplexities for 4g, yahoo-4g, yahoo-5g, sarv, and plm are
126.99, 202.14, 196.76, 128.38, 280.64, respectively.
Note that the two structured 5-gram LMs are trained on a small
subset of the data used for training the word 4-gram, with the parser LM on an even smaller subset, but still the perplexity from the almost-parsing LM is almost the same as the word 4-gram LM. 

\vspace{-0.05in}
\subsection{Reranking experiments}
\vspace{-0.05in}
For the AE and CE SMT tasks, we set the N-best
list size to 3,000%
\footnote{This nbest list size is selected for efficiency of MT decoding.
We will experiment with decoding and rescoring with larger nbest lists in future work.}.
The word 4-gram is used in search and google, yahoo, sarv,
plm, and wlm LMs are used for computing scores for each N-best hypothesis.
The resulting LM scores are combined in a log-linear
framework with weights optimized on the GALE dev07 test set and tested on
the blind test set, NIST eval06 GALE subset in the 2006 NIST MT
evaluation. Weights
for knowledge sources are optimized on the combined set of NW and WT for the text MT,
and BN and BC for the audio MT, respectively, to reduce the risk of over-fitting and
dependence on accurate genre detections if doing optimization on the four genres, NW, WT, BN,
and BC, separately. 
Note that for all of the experiments in this paper, 
we use BLEU scores based on one reference to measure MT performance.

\begin{table}[htb]
\caption{{\small Effect of using Yahoo LMs with different training and rescoring schemes. BLEU sores [\%] are measured for the GALE dev07 and eval06 Arabic-English BN and BC data. The baseline is to use the word 4-gram in search and the Google 5-gram count-LM for reranking.}}
\savespacebeforetable
\label{yahoo-lm}
\begin{center}
\begin{tabular}{|l||c|c||c|c|} \hline
{\bf LMs}                             	& \multicolumn{2}{|c||}{\bf dev07} & \multicolumn{2}{|c|}{\bf 2006-nist/gale} \\
					& \multicolumn{2}{|c||}{\bf AE BLUE [\%]} & \multicolumn{2}{|c|}{\bf AE BLUE [\%]} \\
					 \cline{2-5}
					& BN & BC & BN & BC \\ \hline \hline
Baseline(4g+google)			&	29.38	&	25.56 &	21.99	&	21.88 \\ \hline
+yahoo-4g (KN)                     	&       29.87	&	25.86 &	22.34	&	22.35 \\
+yahoo-5g (count-lm)                 	&	30.06	&     	25.46 &	22.31	&	21.39 \\ \hline
+yahoo-4g + Yahoo-5g 			&	29.65	&	25.98 &	22.19	&	21.07 \\
(dynamic-interp, 0.6)			&		&		& & \\
+yahoo-4g + yahoo-5g 			&	29.57	&	26.07 &	22.59	&	21.72 \\
(static-interp)				& & & & \\
+yahoo-4g + yahoo-5g 		&	29.93	&	25.91 &	22.42	&	22.45 \\
(log-linear)				& & & & \\ \hline
\end{tabular}
\end{center}
\end{table}

In the first experiment, we trained three variations of LMs using the
Yahoo N-grams, i.e.,  a 4-gram LM using modified KN smoothing after
extrapolation (yahoo-4g) and its 5-gram version (yahoo-5g), and a
5-gram count-LM using all of the Yahoo N-grams, denoted yahoo-5g
(count-lm). The effect of using them for reranking dev07 and eval06
nbests is shown in Table~\ref{yahoo-lm}.  Note that on dev07 BN, the
5-gram count-LM using all of the Yahoo N-grams produced greater
improvement than the modified KN smoothed Yahoo 4-gram, but the
observation is reversed on BC. On the blind test set eval06, in fact
yahoo-4g outperformed the 5-gram count-LM on both BN and BC. We
compared the rescoring schemes for the modified KN smoothed Yahoo
4-gram and 5-gram, by using a dynamic (with weight as 0.6 for yahoo-4g) or static interpolation of them
during rescoring, or using them in the log-linear framework. The
picture is again different between BN and BC, on dev07 and eval06. On
dev07 BN, the log-linear scheme yields the best improvement while on
BC the winner is the static interpolation approach. On eval06, the
static interpolation approach works best on BN but the log-linear
scheme is best for BC. Since the gain in BLEU comes mostly from
yahoo-4g (KN), for efficiency of reranking, we used yahoo-4g (KN) for
reranking in the following experiments.

\begin{table}[htb]
\caption{{\small Effect of using the almost-parsing LM (sarv) for N-best reranking, after adding google and yahoo LM scores. BLEU scores [\%] are measured for the GALE dev07 and eval06 Arabic-English BN and BC data.  Unsupervised adaptation is compared to static interpolation of component sarv LMs.}} 
\savespacebeforetable
\label{sarv-lm-bleu}
\begin{center}
\begin{tabular}{|l||c|c||c|c|} \hline
{\bf LMs}                             	& \multicolumn{2}{|c||}{\bf dev07} & \multicolumn{2}{|c|}{\bf 2006-nist/gale} \\
					& \multicolumn{2}{|c||}{\bf AE BLUE [\%]} & \multicolumn{2}{|c|}{\bf AE BLUE [\%]} \\
					\cline{2-5}
					& BN & BC & BN & BC \\ \hline \hline 
(1): 4g + google + yahoo	&	29.87	&	25.86	&	22.34	&	22.35 \\ \hline
(2): (1) + dynamically    	&	30.59	&	25.45	&	22.27	&	22.52  \\
interpolated sarv				& & & & \\
(3): (1) + statically    	&	30.12	&	25.97	&	22.54	&	22.39 \\ 
interpolated sarv				& & & & \\ \hline
\end{tabular}
\end{center}
\end{table}

To further explore the advantages of the almost-parsing LM
in capturing domain-specific grammatical features and word use, we
investigated the effect of unsupervised adaptation. Note
that we trained component SuperARV LMs after clustering sources in its
training data and when using static interpolation, all of the component LMs
are linearly interpolated with weights optimized on a LM tuning set to minimize its perplexity. 
We compared the reranking effect from
statically or dynamically interpolating these component LMs. For
dynamic interpolation, we computed the linear interpolation weights of
component LMs by optimizing perplexities on the 1-best decoding
hypothesis for each sentence. The results are shown in Table \ref{sarv-lm-bleu}. Unsupervised adaptation produced improvement on dev07 BN but not on BC. On eval06, the observation is reversed. Still, one of the interpolation approaches yields improvement on BLEU over the baseline, 4g+google+yahoo. In future work, we will investigate possible factors contributing to this difference on performance of unsupervised adaptation, as well as effective adaptation approaches for the structured LMs.

\begin{table}[htb]
\caption{{\small Effect of LM reranking on the dev07 test set for all the genres for AE.}}
\savespacebeforetable
\label{AE-dev07-bleu}
\begin{center}
\begin{tabular}{|l||c|c||c|c|} \hline
{\bf LMs}                             & \multicolumn{4}{|c|}{\bf dev07 AE BLEU [\%]} \\ \cline{2-5}
                           &	NW     & WT  &   BN  &   BC  \\ \hline \hline
(1) 4gram                  &	31.20    &	22.99   &	28.72   &	25.22 \\ \hline
(2) 4g+google              &	31.45    &	23.66   &	29.38   &	25.56 \\
(3) 4g+yahoo               &	31.20    &	22.99   &	29.46   &	25.78 \\
(4) 4g+google+yahoo        &	31.45    &	23.65   &	29.87   &	25.86 \\ \hline
(5): (4)+sarv              &	31.71    &	23.59   &	30.59   &	25.45 \\
(6): (4)+sarv+plm          &	32.15    &	23.99   &	{\bf 31.12}   &	26.30 \\ \hline
(7): (4)+sarv+plm+wlm      &	{\bf 32.18}    &	{\bf 24.07}   &	31.04   &	{\bf 26.48} \\ \hline
\end{tabular}
\end{center}
\end{table}

\begin{table}[htb]
\caption{{\small Effect of LM reranking on the blind test set, 2006 NIST MT evaluation Arabic-English GALE data.}}
\savespacebeforetable
\label{AE-eval06-bleu} 
\begin{center}  
\begin{tabular}{|l||c|c||c|c|} \hline 
{\bf LMs}                             & \multicolumn{4}{|c|}{\bf 2006-nist/gale AE BLEU [\%]} \\ \cline{2-5} 
                           &	NW     & WT  &   BN  &   BC  \\ \hline \hline 
(1) 4gram                  &	27.36    &	15.59   &	21.67   &	21.58 \\ \hline
(2) 4g+google              &	28.09    &	16.17   &	21.99   &	21.88 \\
(3) 4g+yahoo               &	27.33    &	15.59   &	22.33   &	22.15 \\
(4) 4g+google+yahoo        &	28.09    &	16.17   &	22.34   &	22.35 \\ \hline
(5): (4)+sarv              &	28.01    &	16.26   &	22.27   &	22.52 \\
(6): (4)+sarv+plm          &	{\bf 28.22}    &	16.16   &	23.01   &	{\bf 23.14} \\ \hline
(7): (4)+sarv+plm+wlm      &	28.11    &	{\bf 16.43}   &	{\bf 23.02}   &	23.05 \\ \hline
\end{tabular}
\end{center} 
\end{table} 

\begin{table}[htb]
\caption{{\small Effect of LM reranking on the dev07 test set for all the genres for CE.}}
\savespacebeforetable
\label{CE-dev07-bleu}
\begin{center}
\begin{tabular}{|l||c|c||c|c|} \hline
{\bf LMs}                             & \multicolumn{4}{|c|}{\bf dev07 CE BLEU [\%]} \\ \cline{2-5}
                           &	NW	& WT  		&   BN  	&   BC  \\ \hline \hline
(1) 4gram                  &	16.72    &	14.63   &	19.62   &	16.34  \\ \hline 
(2) 4g+google              &	17.16    &	14.62   &	20.00   &	16.47  \\ 
(3) 4g+yahoo               &	17.16    &	14.59   &	19.87   &	16.40  \\ 
(4) 4g+google+yahoo        &	17.24    &	14.72   &	20.17   &	16.37  \\ \hline
(5): (4)+sarv              &	17.40    &	14.89   &	20.44   &	16.27  \\ 
(6): (4)+sarv+plm          &	17.51    &	15.11   &	20.56   &	16.70  \\ \hline
(7): (4)+sarv+plm+wlm      &	{\bf 17.51}    &	{\bf 15.23}   &	{\bf 20.77}   &	{\bf 16.81}  \\ \hline
\end{tabular}
\end{center}
\end{table}

\begin{table}[htb]
\caption{{\small Effect of LM reranking on the blind test set, 2006 NIST MT evaluation Chinese-English GALE data.}}
\savespacebeforetable
\label{CE-eval06-bleu}
\begin{center}
\begin{tabular}{|l||c|c||c|c|} \hline
{\bf LMs}                             & \multicolumn{4}{|c|}{\bf 2006-nist/gale CE BLEU [\%]} \\ \cline{2-5}
                           & NW     & WT  &   BN  &   BC  \\ \hline \hline
(1): 4gram                  &   17.71 & 14.16 & 17.17 & 15.51 \\ \hline
(2): 4g+google              &     18.03 & 14.84 & 17.16 & 15.51 \\ 
(3): 4g+yahoo               &    18.28 & 14.73 & 17.44 & 15.33 \\ 
(4): 4g+google+yahoo        &      18.45 & 14.91 & 17.16 & 15.28 \\ \hline
(5): (4)+sarv               &   {\bf 18.73} & {\bf 15.50} & 17.51 & 15.55 \\ 
(6): (4)+sarv+plm           &   18.61 & 15.09 & 17.56 & 15.95 \\ \hline
(7): (4)+sarv+plm+wlm       &   18.72 & 15.04 & {\bf 17.75} & {\bf 16.05} \\ \hline
\end{tabular}
\end{center}
\end{table}

Table~\ref{AE-dev07-bleu}, \ref{AE-eval06-bleu}, \ref{CE-dev07-bleu},
\ref{CE-eval06-bleu} present the BLEU scores from the baseline (i.e.,
no-reranking) and using various LMs for reranking on dev07 and eval06
for both Arabic-English and Chinese-English.  On the Arabic-English
dev07 test set, almost all of the LMs improve BLEU scores
incrementally over the baseline 4-gram (or at least no degradation)
and the combinations of all of the LMs for reranking yield the best
improvement on  all of the four genres with the exception of the
BBN-web-lm on BN. The absolute improvement in BLEU scores ranges from
0.98\% on NW, 1.08\% on WT,  2.4\% on BN, and 1.26\% on BC.  On the AE
blind test set, 2006-nist/gale AE, the absolute improvement in BLEU
scores from the combination of google and yahoo LMs ranges from 0.73\%
on NW,  0.58\% on WT, 0.67\% on BN, and 0.77\% on BC.  The combination
of the two structured language models produced absolute improvement in
BLEU scores of 0.67\% on BN and 0.79\% on BC.  Adding the BBN-web-lm
for reranking further helps improving the BLEU score on WT by 0.27\%
absolutely.  The absolute gain from LM reranking on the AE blind test
sets ranges from 0.86\% on NW, 0.84\% on WT, 1.35\% on BN, and 1.56\%
on BC.
   
On the Chinese-English dev07 test set, the combination of all LMs
yields the best improvement in BLEU scores on all of the four genres,
ranging from 0.79\% on NW, 0.6\% on WT, 1.15\% on BN, and 0.47\% on
BC.     On the CE blind test set, the absolute gain from the
combination of google and yahoo LMs ranges from  0.74\% on NW, 0.75\%
on WT, no improvement on BN and BC (although the yahoo LM improves BN by
0.27\%).     Adding the almost-parsing LM improves the BLEU scores for
each genre, up to 0.59\% absolutely on WT. On the other hand, the
parser LM helps especially for the BC genre with 0.4\% absolutely. The
BBN-web-lm also produced some small gains on all genres except WT.
The combination of all of them tends to produce the best BLEU
performance, except for the WT genre. The absolute gain from LM
reranking on the CE blind test sets ranges from 1.02\% on NW, 1.34\%
on WT, 0.58\% on BN, and 0.54\% on BC.

In all, for AE, these LMs for reranking help more on BN and BC; for
CE, they help more on NW and WT.  There are some fluctuations in BLEU
scores when adding a new LM for reranking 
but in all, the combination
of all these structured LMs and Yahoo and Google LMs tends to produce
the best improvement, up to 1.6\% absolutely on the blind test set. 
It
is also noticeable that the patterns of gains from LMs are partially
different or reversed between BN and BC, partly due to the significant
difference between the two genres and the differences between the training
data of these LMs and language features captured by them. Also, the patterns of gains are
different between dev07 and eval06 in some cases, indicating there
might be significant difference between dev07 and eval06 and the
weight optimization approach could be improved to reduce the chance of
getting trapped in a local optimum.

In future work, we will investigate genre-specific grammatical
phenomena (e.g., speech disfluency for BC and the unstructured
characteristics for WT) for structured LMs and effectiveness of
adaptation approaches on them. We will also investigate more effective
and robust approaches for combining multiple LMs for reranking as well
as efficient approaches to employ more sophisticated LMs in search.

\footnotesize

\section{Acknowledgments}
We thank Richard Zens and Oliver Bender from RWTH Aachen University for making the MT training corpora and alignments
available.
This material is based upon work supported by
the Defense Advanced Research Projects Agency (DARPA) under Contract
No. HR0011-06-C-0023 (approved for public release, distribution unlimited).
Any opinions, findings and conclusions or
recommendations expressed in this material are those of the authors
and do not necessarily reflect the views of DARPA.

\scriptsize

\bibliographystyle{ieee-shortnames}
\bibliography{reference,all-short}

\newpage
\normalsize
\section*{Correction}
This version of the paper corrects an error in the caption and legend to Figure~\ref{fig:kncounts}.  
The original (as published in ASRU 2007) gave an incorrect functional form for the values plotted on the y-axis,
which was therefore inconsistent with Equation~(\ref{eq:kncounts}).
The plot itself is unchanged.

\end{document}